\documentclass[letterpaper, 10 pt, conference]{ieeeconf}  

\IEEEoverridecommandlockouts                              

\usepackage{graphics} 
\usepackage{graphicx} 
\usepackage{epsfig} 
\usepackage{amsmath} 
\usepackage{amssymb}  
\usepackage{booktabs} 
\usepackage{multirow} 
\usepackage{algorithm} 
\usepackage{algpseudocode} 
\usepackage{url} 
\usepackage{cite} 
\usepackage{color}
\usepackage{balance}
\usepackage{placeins}
\newcommand{\methodname}{\textbf{ProbeFlow}} 

\title{\LARGE \bf
\methodname{}: Training-Free Adaptive Flow Matching for Vision-Language-Action Models
}

\author{
    Zhou Fang$^{1, 2}$, 
    Jiaqi Wang$^{1, 2}$, 
    Wenzhuo Li$^{1, 2}$, 
    Qiongfeng Shi$^3$,
    Yi Zhou$^{1, 2*}$, 
     \\ [1ex]
    $^1$ School of Computer Science and Engineering, Southeast University, China \\
    $^2$ Key Laboratory of New Generation Artificial Intelligence Technology and Its Interdisciplinary \\ 
    Applications, Ministry of Education, China \\
    $^3$ School of Electronic Science \& Engineering, Southeast University, China \\ [1ex]
    {\tt\small \{zfan, jiaqi.wang, wenzhuoli, qiongfeng\}@seu.edu.cn \quad yizhou.szcn@gmail.com}
    \thanks{$^*$ Corresponding author.}
}

\begin{document}

\maketitle
\thispagestyle{empty}
\pagestyle{empty}


\begin{abstract}
Vision-Language-Action (VLA) policies increasingly use Flow Matching (FM) action heads to generate action chunks as continuous trajectories. At inference, an FM head solves an ordinary differential equation (ODE): starting from a prior action latent, it repeatedly queries a learned velocity field to advance the state toward the decoded action chunk. Because each query depends on the preceding state, this iterative decoding can impose substantial sequential action-head cost at deployment. We propose \methodname{}, a training-free framework that routes each action chunk to an integration schedule using a Lookahead Consistency Probe. Optional oscillation-conditioned (OSC) threshold adaptation reuses the same velocity evaluations and can trade part of the acceleration for potentially better task performance. On MetaWorld, the fixed-threshold configuration uses 2.60 scheduled steps (2.92 action-head evaluations) and reports 15.9 ms action-head latency, versus 235.7 ms for 50-step Euler, while obtaining 83.2\% success versus 82.5\%. On LIBERO, it uses 4.53 scheduled steps (5.09 action-head evaluations), reports 32.7 ms, and obtains 88.7\% success, compared with 278.7 ms and 92.5\% for 50-step Euler. We also evaluate OSC on three real-world manipulation tasks with 50 trials per method and task. These evaluations characterize action-decoding compute--performance trade-offs without additional training.
\end{abstract}


\section{INTRODUCTION}

Vision-Language-Action (VLA) models~\cite{brohan2022rt,zitkovich2023rt,kim2024openvla} map visual observations and language instructions to robot actions. Flow Matching (FM)~\cite{lipman2022flow} is a continuous action-head formulation that represents an action chunk through a conditional velocity field. At deployment, decoding solves an ordinary differential equation (ODE): beginning from a prior action latent, the policy follows the learned velocity field over an integration interval until it reaches the action chunk. Numerical solvers approximate this path through a sequence of field evaluations, and each evaluation depends on the state produced by the preceding one.

This sequential dependence can make action decoding a material deployment cost. Much VLA acceleration work focuses on the Vision-Language Model (VLM) backbone~\cite{wang2025bitvla, yang2025efficientvla, yue2024deer}, while an FM action head still applies repeated field evaluations to produce one action chunk. A fixed solver must allocate the same evaluation budget to easy transit segments and locally changing segments, despite their potentially different integration requirements. We therefore propose \methodname{} with two complementary, training-free routing components. The \textit{Lookahead Consistency Probe} compares two velocity evaluations to form a local consistency signal and maps it to a discrete integration budget. Optional \textit{oscillation-conditioned} (OSC) adaptation reuses these evaluations to adjust the routing threshold from a sign-change statistic. The resulting router selects a minimum two-segment update under high consistency and a denser Euler schedule otherwise; OSC can choose a more conservative schedule for relevant chunks.

\begin{figure*}[t]
    \centering
    \includegraphics[width=0.95\textwidth]{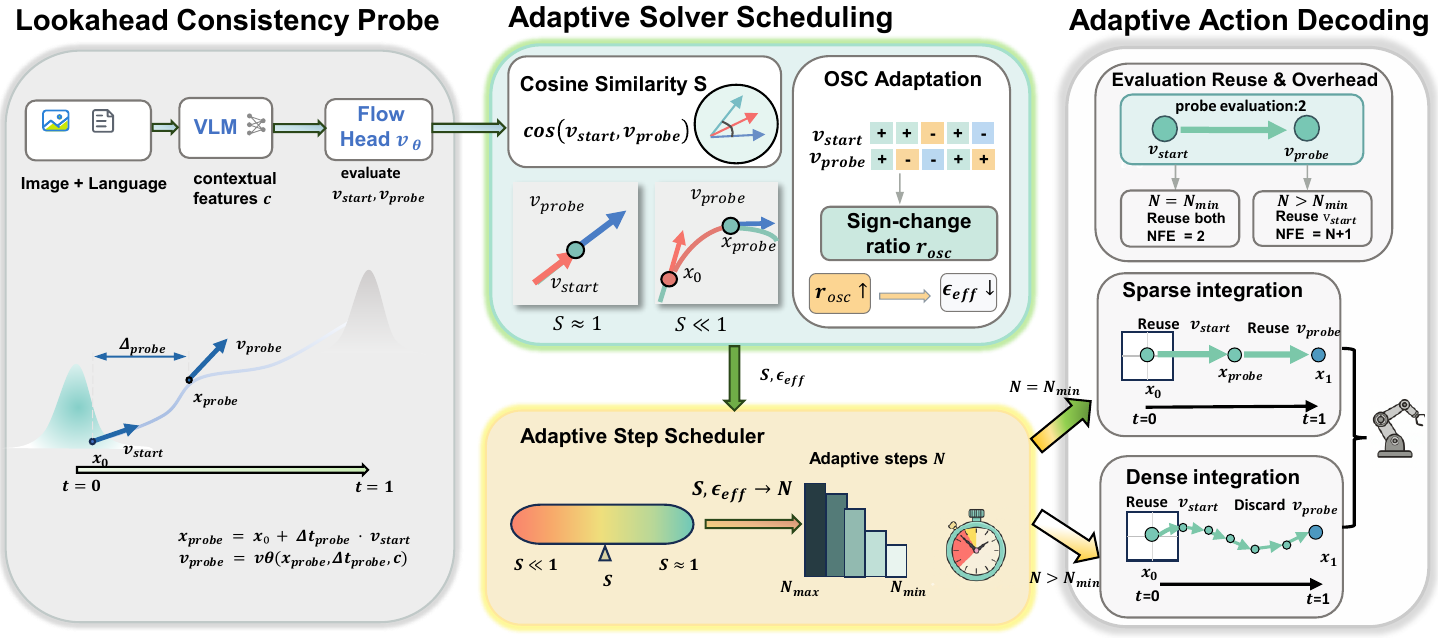}
    \caption{Overview of the proposed \methodname{} framework. \textbf{Left:} A conditional action flow is integrated from a prior state to an action chunk. \textbf{Middle:} The Lookahead Consistency Probe compares initial and lookahead velocity directions through cosine similarity $\mathcal{S}$, which the scheduler maps to an adaptive step count $N$ under a deployment latency budget; optional OSC maps a sign-change ratio to an effective threshold. \textbf{Right:} High directional consistency ($\mathcal{S} \approx 1$) selects a coarse schedule that reuses the two probe evaluations; lower consistency selects a denser schedule.}
    \label{fig:backbone}
\end{figure*}

In summary, our main contributions are as follows:
\begin{itemize}
    \item We propose \methodname{}, a training-free adaptive solver that targets the action-decoding cost of continuous generative policies.
    \item We introduce two complementary, training-free routing mechanisms: a \textit{Lookahead Consistency Probe} that computes cosine similarity between initial and lookahead velocity vectors, and an optional \textit{oscillation-conditioned} (OSC) threshold mechanism that reuses these vectors to adjust routing conservatism online, exposing an accuracy--acceleration trade-off.
    \item Evaluations on MetaWorld and LIBERO measure the resulting compute--performance operating points for both fixed-threshold and OSC configurations. On MetaWorld, the fixed-threshold configuration reports 2.60 scheduled steps (2.92 NFE), 15.9 ms action-head latency, and 83.2\% success, compared with 235.7 ms and 82.5\% for 50-step Euler. We additionally report real-world action-head measurements and task outcomes under a fixed setup.
    \item We open-source the implementation and evaluation code at \url{https://github.com/zfan00666/probeflow}.
\end{itemize}
An overview of the proposed framework is illustrated in Fig.~\ref{fig:backbone}.

\section{RELATED WORK}

\subsection{Flow Matching in Vision-Language-Action Models}
Flow Matching (FM)~\cite{lipman2022flow} has been used as a continuous action-head formulation for VLA policies such as $\pi_0$~\cite{black2024pi_0} and Evo-1~\cite{lin2025evo}. It represents a conditional action distribution through a time-dependent vector field and decodes action chunks through repeated vector-field evaluations. This iterative inference path motivates action-decoding acceleration.

\subsection{Efficiency Optimizations in VLA Models}
To mitigate the computational overhead of VLA models~\cite{yu2025survey}, prior work has optimized vision-language backbones through quantization~\cite{wang2025bitvla} and token compression~\cite{yang2025efficientvla}. These approaches reduce the cost of perceptual and language processing and are complementary to reducing the iterative computation of a generative action head.

Adaptive action decoding has also been studied for imitation learning. AdaFlow~\cite{hu2024adaflow} trains an auxiliary variance estimator to adjust the integration schedule of a flow-based policy. In contrast, \methodname{} uses a training-free comparison of two action-head velocity evaluations to route the current action chunk to an integration schedule. It requires no additional optimization or parameter updates to the pre-trained policy.

\subsection{Advanced ODE Solvers in Generative Models}
Advanced ODE/SDE solvers accelerate continuous-time generative models through higher-order updates or adaptive numerical step sizes. DPM-Solver and DPM-Solver++~\cite{lu2022dpm, lu2025dpm} exploit the structure of diffusion ODEs, while RK45~\cite{dormand1980family} adapts numerical step sizes using embedded error estimates. Heun's method uses a predictor--corrector update with velocities evaluated at the start and provisional endpoint of each step~\cite{atkinson2009numerical}. For a directly parameterized Flow Matching velocity field, the corresponding fixed-step multistep reference is AB2~\cite{atkinson2009numerical}. In contrast, \methodname{} adaptively allocates an Euler integration budget from an online directional-consistency signal.

\section{METHODOLOGY}

\subsection{Lookahead Consistency Probe}
\methodname{} replaces the fixed integration budget with an online routing decision. Its one-shot \textit{Lookahead Consistency Probe} estimates local directional consistency from two action-head evaluations. The resulting score selects either a coarse update or a denser schedule. The probe uses two evaluations already compatible with the action-decoding path and does not require derivatives, an auxiliary predictor, or additional training.

Given the initial noise state $\boldsymbol{x}_0$ at integration time $t=0$, we first compute the initial velocity vector using the trained flow matching action head:
\begin{equation}
    \boldsymbol{v}_{\mathrm{start}} = \boldsymbol{v}_\theta(\boldsymbol{x}_0, 0, \boldsymbol{c}).
\end{equation}
The probe takes a lookahead step of size $\Delta t_{\mathrm{probe}}$ (e.g., $\Delta t_{\mathrm{probe}} = 0.5$) to project a future state:
\begin{equation}
    \boldsymbol{x}_{\mathrm{probe}} = \boldsymbol{x}_0 + \boldsymbol{v}_{\mathrm{start}} \cdot \Delta t_{\mathrm{probe}}.
\end{equation}
Subsequently, we evaluate the vector field at this newly probed state:
\begin{equation}
    \boldsymbol{v}_{\mathrm{probe}} = \boldsymbol{v}_{\theta}(\boldsymbol{x}_{\mathrm{probe}}, \Delta t_{\mathrm{probe}}, \boldsymbol{c}).
\end{equation}
We measure the directional consistency between the initial and lookahead velocity vectors using cosine similarity:
\begin{equation}
    \mathcal{S} = \frac{\boldsymbol{v}_{\mathrm{start}}^\top \boldsymbol{v}_{\mathrm{probe}}}{\| \boldsymbol{v}_{\mathrm{start}} \|_2 \| \boldsymbol{v}_{\mathrm{probe}} \|_2}.
\end{equation}
Geometrically, $\mathcal{S}$ is the cosine of the angle $\phi$ between the initial direction and the field evaluated at the probe state. A value near one indicates that the two evaluations agree directionally over the selected interval, whereas a lower value indicates greater directional change. We use this signal as an empirical routing feature: high consistency selects fewer integration steps, while lower consistency selects a denser schedule. The probe therefore provides a local estimate of how rapidly the velocity field changes along the integration path. When the two evaluated velocities remain directionally aligned, the local trajectory is approximately smooth over the probe interval, and the two-segment update can approximate the endpoint with a small evaluation budget. A larger directional change indicates stronger variation of the field along the path and motivates a denser Euler schedule. This interpretation is local and empirical: the cosine score serves as a routing signal rather than a formal numerical error estimator. 

\subsection{Dynamic Step Scheduling and Integration}
Given the directional-consistency score $\mathcal{S}$, \methodname{} maps this continuous signal to a discrete number of integration steps $N$ for the current action chunk:
\begin{equation}
    N = \text{clip} \left( N_{\min} + \lfloor \frac{1 - \mathcal{S}}{\epsilon} \rfloor \times \Delta N, \; N_{\min}, \; N_{\max} \right),
\end{equation}
where $\epsilon$ is a sensitivity hyperparameter defining the tolerance for directional deviation, $\Delta N$ denotes the discrete step increment size, and the bounds $N_{\min}$ and $N_{\max}$ constrain the computational budget according to the deployment latency target. The minimum schedule is fixed as $N_{\min}=2$ in our implementation and experiments.

\methodname{} uses conditional routing to reuse the probe evaluations whenever possible. When high directional consistency selects the minimum schedule ($N=N_{\min}=2$), the two probe evaluations form a two-segment update to the final predicted action chunk $\hat{\boldsymbol{a}}$:
\begin{equation}
    \hat{\boldsymbol{a}} \equiv \boldsymbol{x}_{t=1} = \boldsymbol{x}_{\mathrm{probe}} + \boldsymbol{v}_{\mathrm{probe}} \cdot (1 - \Delta t_{\mathrm{probe}}).
\end{equation}
This branch uses the same two action-head evaluations required by the probe; no separate solver evaluation is introduced after the lookahead evaluation. When lower directional consistency selects $N > N_{\min}$, the method instead reuses $\boldsymbol{v}_{\mathrm{start}}$ to initialize a denser Euler schedule:
\begin{equation}
    \boldsymbol{x}_{\Delta t} = \boldsymbol{x}_0 + \boldsymbol{v}_{\mathrm{start}} \cdot \Delta t, \quad \text{where } \Delta t = 1/N.
\end{equation}
The lookahead evaluation $\boldsymbol{v}_{\mathrm{probe}}$ is not used by this Euler branch, while the initial evaluation $\boldsymbol{v}_{\mathrm{start}}$ is reused. Relative to a fixed $N$-step Euler solver, this branch therefore adds one lookahead action-head evaluation. The complete execution pipeline is summarized in Algorithm~\ref{alg:adaflow}.

The computational cost of \methodname{} depends on the selected routing branch. Every action chunk first incurs the two evaluations required by the probe. If the minimum schedule is selected, these evaluations are reused directly in the final two-segment update, so the total NFE remains two. If a denser schedule is selected, the initial velocity evaluation is reused to initialize the Euler trajectory, while the lookahead evaluation supplies the routing decision and is not reused by the Euler updates. The resulting cost is therefore $N+1$ action-head evaluations for a scheduled count $N>2$. This distinction explains why measured latency is not determined by the scheduled step count alone and motivates reporting both scheduled $N$ and NFE.

The complete execution pipeline is summarized in Algorithm~\ref{alg:adaflow}.

\subsection{Oscillation-Conditioned Scheduling}
\label{sec:oscillation_conditioned}

The fixed threshold $\epsilon$ provides a simple global control of the speed--accuracy trade-off. We introduce an optional \textit{oscillation-conditioned} (OSC) threshold mechanism that computes an additional sign-change statistic from the velocity vectors already evaluated by the lookahead probe and uses it to modulate the threshold online, without introducing an auxiliary network or changing the probe horizon.

Given the initial and lookahead velocities, we measure the fraction of action dimensions whose signs change beyond a small deadband $\delta$:
\begin{equation}
    \begin{aligned}
    r_{\mathrm{osc}}
    &= \max_b \frac{1}{D} \sum_{j=1}^{D}
    \mathbb{I}\left[\operatorname{sign}(v_{\mathrm{start},b,j})
    \ne \operatorname{sign}(v_{\mathrm{probe},b,j})\right] \\
    &\quad \cdot \mathbb{I}\left[
    \max\left(|v_{\mathrm{start},b,j}|, |v_{\mathrm{probe},b,j}|\right) > \delta
    \right].
    \end{aligned}
\end{equation}
where $D$ is the flattened action dimension and the maximum is taken over the batch. We normalize this sign-change statistic using a reference level $r_{\mathrm{ref}}$:
\begin{equation}
    \tilde r_{\mathrm{osc}} = \operatorname{clip}\left(\frac{r_{\mathrm{osc}}}{r_{\mathrm{ref}}},0,1\right).
\end{equation}

The effective threshold is then set to
\begin{equation}
    \begin{aligned}
    \epsilon_{\mathrm{eff}}
    = \operatorname{clip}\bigl(&\epsilon_0 [s_{\mathrm{easy}}
    - (s_{\mathrm{easy}}-s_{\mathrm{hard}})\tilde r_{\mathrm{osc}}], \\
    &\epsilon_0 s_{\min},\ \epsilon_0 s_{\max}\bigr).
    \end{aligned}
\end{equation}
where $\epsilon_0$ is the base threshold. In the current configuration, $s_{\mathrm{easy}}>1$ relaxes the threshold for lower values of the sign-change statistic, while $s_{\mathrm{hard}}<1$ tightens it for higher values. The resulting $\epsilon_{\mathrm{eff}}$ replaces $\epsilon$ in the step scheduler. Thus, OSC maps this additional probe statistic to a more or less conservative integration budget.

OSC is a threshold-adaptation mechanism rather than a probe-horizon adaptation mechanism: $\Delta t_{\mathrm{probe}}$ remains fixed in the current implementation. It is also training-free and adds only lightweight statistics over already computed velocity vectors.

\begin{algorithm}[t]
\caption{\methodname{} Inference Pipeline}
\label{alg:adaflow}
\begin{algorithmic}[1]
\Require Trained model $\boldsymbol{v}_\theta$, Initial noise $\boldsymbol{x}_0$, Condition $\boldsymbol{c}$, Base threshold $\epsilon_0$, optional OSC parameters, $\Delta t_{\mathrm{probe}}$, Bounds $N_{\min}, N_{\max}$, Increment $\Delta N$

\State $\boldsymbol{v}_{\mathrm{start}} \leftarrow \boldsymbol{v}_\theta(\boldsymbol{x}_0, 0, \boldsymbol{c})$
\State $\boldsymbol{x}_{\mathrm{probe}} \leftarrow \boldsymbol{x}_0 + \boldsymbol{v}_{\mathrm{start}} \cdot \Delta t_{\mathrm{probe}}$
\State $\boldsymbol{v}_{\mathrm{probe}} \leftarrow \boldsymbol{v}_\theta(\boldsymbol{x}_{\mathrm{probe}}, \Delta t_{\mathrm{probe}}, \boldsymbol{c})$
\State $\mathcal{S} \leftarrow \frac{\boldsymbol{v}_{\mathrm{start}}^\top \boldsymbol{v}_{\mathrm{probe}}}{\| \boldsymbol{v}_{\mathrm{start}} \|_2 \| \boldsymbol{v}_{\mathrm{probe}} \|_2}$

\State $\epsilon_{\mathrm{eff}} \leftarrow \mathrm{OSC}(\boldsymbol{v}_{\mathrm{start}}, \boldsymbol{v}_{\mathrm{probe}}, \epsilon_0)$ if OSC is enabled; otherwise $\epsilon_{\mathrm{eff}} \leftarrow \epsilon_0$

\State $N \leftarrow \mathrm{clip} \left( N_{\min} + \Delta N \times \lfloor \frac{1 - \mathcal{S}}{\epsilon_{\mathrm{eff}}} \rfloor, N_{\min}, N_{\max} \right)$

\If{$N = N_{\min}$}
    \State $\hat{\boldsymbol{a}} \leftarrow \boldsymbol{x}_{\mathrm{probe}} + \boldsymbol{v}_{\mathrm{probe}} \cdot (1 - \Delta t_{\mathrm{probe}})$
\Else
    \State $\Delta t \leftarrow 1.0 / N$
    \State $\boldsymbol{x}_t \leftarrow \boldsymbol{x}_0 + \boldsymbol{v}_{\mathrm{start}} \cdot \Delta t$
    \State $t \leftarrow \Delta t$
    \For{$i = 1$ \textbf{to} $N - 1$}
        \State $\boldsymbol{v} \leftarrow \boldsymbol{v}_\theta(\boldsymbol{x}_t, t, \boldsymbol{c})$
        \State $\boldsymbol{x}_t \leftarrow \boldsymbol{x}_t + \boldsymbol{v} \cdot \Delta t$
        \State $t \leftarrow t + \Delta t$
    \EndFor
    \State $\hat{\boldsymbol{a}} \leftarrow \boldsymbol{x}_t$
\EndIf
\State \Return $\hat{\boldsymbol{a}}$
\end{algorithmic}
\end{algorithm}

\section{EXPERIMENTS}

We evaluate \methodname{} along three axes: 
\textbf{1) Efficiency--performance trade-off:} How does adaptive action decoding compare with fixed-step solvers at the evaluated operating points? 
\textbf{2) Routing behavior:} Does the lookahead signal induce different integration budgets across generated action flows? 
\textbf{3) Sensitivity:} How do the threshold and probe horizon affect the measured compute--performance trade-off?

\subsection{Experimental Setup}

\noindent \textbf{Model Architecture.} 
We use the \textit{Evo-1}~\cite{lin2025evo} architecture with a frozen InternVL3-1B~\cite{zhu2025internvl3} visual backbone. The action head is a conditional Flow Matching network implemented via an 8-layer Diffusion Transformer (DiT~\cite{peebles2023scalable}) with a 1024 hidden dimension, trained to predict trajectories of horizon $T=50$.

\noindent \textbf{OSC Configuration.} Unless otherwise specified, the optional OSC mechanism uses $r_{\mathrm{ref}}=0.02$, $s_{\mathrm{easy}}=1.50$, $s_{\mathrm{hard}}=0.75$, and scale bounds $[s_{\min},s_{\max}]=[0.60,1.60]$. These parameters modulate the effective threshold around the base value and do not change $\Delta t_{\mathrm{probe}}$, $N_{\min}$, or $N_{\max}$. The fixed-threshold variant disables OSC and uses $\epsilon_{\mathrm{eff}}=\epsilon_0$.

\noindent \textbf{Benchmarks.}
We evaluate on MetaWorld (MT50)~\cite{yu2020meta}, comprising 50 manipulation tasks, and LIBERO~\cite{liu2023libero}, which contains multi-stage, language-conditioned tasks. The fixed-threshold and baseline rows in Tables~\ref{tab:metaworld_results}, \ref{tab:libero_results}, and \ref{tab:ablation_epsilon} report 5 evaluation seeds with 10 evaluation episodes per task and seed. The OSC rows in Tables~\ref{tab:metaworld_results}, \ref{tab:libero_results}, and \ref{tab:osc_results} aggregate 3 evaluation seeds. The real-world study uses one evaluation seed and reports trial-level Wilson intervals separately.

\noindent \textbf{Baselines \& Implementation Details.}
We compare \methodname{} against standard fixed-step Euler integration solvers with $N \in \{2,3,10,20,50\}$ on MetaWorld and $N \in \{4,5,10,20,50\}$ on LIBERO. Furthermore, to evaluate high-order acceleration methods, we employ the classical second-order Adams-Bashforth (AB2~\cite{atkinson2009numerical}) method and fixed-step Heun baselines. For the DPM-Solver-inspired multistep reference, we use uniform-time velocity extrapolation (AB2), without the diffusion-specific parameterization and noise-schedule transformation of the original solver. This comparison evaluates a direct velocity-space adaptation, not the full family of DPM-Solver variants. We also include the classic adaptive-step RK45 solver and AdaFlow~\cite{hu2024adaflow} as baselines. AdaFlow uses a learned variance estimator to adapt the integration schedule, whereas \methodname{} performs this adaptation without auxiliary training. Evaluation is performed on a single consumer-grade NVIDIA RTX 4090 GPU.

No task-specific fine-tuning is applied during evaluation. For the fixed-threshold \methodname{} results, we use one configuration across tasks within each reported benchmark: $N_{\min}=2$, $N_{\max}=20$, $\Delta N=2$, $\epsilon=0.008$, and $\Delta t_{\mathrm{probe}}=0.5$. The probe-horizon ablation changes only $\Delta t_{\mathrm{probe}}$; the remaining settings are held fixed. Here, the scheduled integration count $N$ denotes numerical segments, whereas NFE denotes actual action-head evaluations. In Tables~\ref{tab:metaworld_results} and~\ref{tab:libero_results}, the reported count for Heun is NFE (twice the step count in its row label); for Euler and \methodname{} it is the integration schedule $N$. For fixed Euler, $N$ is also the number of action-head evaluations (NFE). For \methodname{}, the minimum branch uses two probe evaluations, while a branch with scheduled $N>2$ uses $N+1$ action-head evaluations because of the additional lookahead evaluation. The measured fixed-threshold averages are 2.60 scheduled steps and 2.92 NFE on MetaWorld, and 4.53 scheduled steps and 5.09 NFE on LIBERO. 

\subsection{Main Results: Latency Breakdown and Success Rate}
\label{sec:main_results}

We profile the action-head inference pipeline on the MetaWorld benchmark. As detailed in Table~\ref{tab:metaworld_results}, fixed-step solvers exhibit a trade-off between solver latency and manipulation performance. Since the visual encoder is unchanged across methods, the table reports only the flow-solver latency affected by \methodname{}.

\begin{table}[t]
\caption{Action-head latency and success rate on MetaWorld. Fixed-threshold and baseline rows aggregate 5 seeds; the OSC row aggregates 3 seeds. For \methodname{} rows, the parenthesized value is the measured average NFE; all probe evaluations are included in action-head latency. Visual encoding is excluded because \methodname{} does not modify the visual encoder.}
\label{tab:metaworld_results}
\centering
\resizebox{\columnwidth}{!}{%
\begin{tabular}{lccc}
\toprule
\textbf{Method} & \textbf{Reported Count} & \textbf{Flow Solver (ms)} $\downarrow$ & \textbf{Success Rate (\%)} $\uparrow$ \\
\midrule
Fixed-Euler ($N=50$) & 50.0 & 235.7  & 82.5 \\
Fixed-Euler ($N=20$) & 20.0 & 99.2  & 80.6 \\
Fixed-Euler ($N=10$) & 10.0 & 53.4  & 81.6 \\
Fixed-Euler ($N=3$)  & 3.0  & 23.7  & 72.4 \\
Fixed-Euler ($N=2$)  & 2.0  & 13.2  & 68.5 \\
RK45                 & 68.9 & 2823.8  & 63.0 \\
AB2 & 10.0 & 65.6  & 78.8 \\
Heun ($N=10$) & 20.0 & 128.78 & 13.6\\
Heun ($N=25$) & 50.0 & 366.09 & 52.0\\
AdaFlow & 2.83 & 20.81 & 83.5\\
\midrule
\methodname{} (fixed $\epsilon$) & \textbf{2.60 (2.92 NFE)} & \textbf{15.9}  & 83.2  \\
\textbf{\methodname{} (OSC)} & 3.84 (4.21 NFE) & 25.8 & \textbf{84.0} \\
\bottomrule
\end{tabular}
}%
\vspace{-0.3cm}
\end{table}

Using a standard 50-step Euler solver, the iterative Flow Matching head consumes \textbf{235.7 ms}. Reducing fixed Euler integration to 10 steps lowers action-head latency to 53.4 ms, with a 0.9 percentage-point reduction in the measured success rate relative to the 50-step setting.

At a similarly small integration budget, Fixed-Euler $N=3$ obtains 72.4\% success, whereas \methodname{} uses an average scheduled $N=2.60$ (2.92 average NFE) and obtains 83.2\%. Relative to the lower-budget Fixed-Euler $N=2$ reference, \methodname{} adds 2.7 ms of action-head latency (15.9 versus 13.2 ms) while improving success by 14.7 percentage points (83.2\% versus 68.5\%). The latter reports \textbf{15.9 ms} action-head latency, which is \textbf{14.8$\times$ lower than the 50-step solver} in this measurement. The NFE includes the additional lookahead evaluation on non-minimum branches; the measured latency includes this cost. In this evaluation, the routing configuration reaches a different operating point from assigning the same small integration schedule to every action chunk.

The advanced-solver baselines illustrate different compute--performance operating points. AB2 uses a fixed 10-step schedule and obtains 78.8\% success. RK45 requires multiple action-head evaluations within an adaptive update and reaches 2823.8 ms in this implementation, exceeding the latency of the other reported solvers. In contrast, \methodname{} uses a single lookahead consistency signal to select its schedule. For schedules above the minimum, this routing decision adds one lookahead action-head evaluation relative to fixed-step Euler; at the minimum schedule, the two probe evaluations are reused in the final two-segment update.

The Heun rows are substantially less competitive in this evaluation, especially on MetaWorld. The predictor--corrector updates alter the intermediate states on which the action head is queried, and this discretization behavior may be less suitable for the pretrained policy's directly parameterized action flow.

Compared with AdaFlow, \methodname{} reports 83.2\% versus 83.5\% MetaWorld success, 15.9 versus 20.81 ms action-head latency, and 2.6 versus 2.83 average steps. On LIBERO, the corresponding reported success rates are 88.7\% and 90.2\%, with action-head latencies of 32.7 and 184.18 ms. This comparison highlights different measured operating points of the two adaptive schedulers: AdaFlow uses a learned variance estimator, while \methodname{} uses a training-free directional-consistency signal.

\subsection{Results on the LIBERO Benchmark}
\label{sec:libero_results}

LIBERO and MetaWorld differ in task construction, rollout length, observations, and policy checkpoints. The $N=50$ baseline reaches 92.5\% success on LIBERO, compared with 82.5\% on MetaWorld; we report this difference as a benchmark-specific observation rather than attributing it to a single factor.

As shown in Table~\ref{tab:libero_results}, the $N=50$ solver requires 278.7 ms for action decoding. The fixed-threshold \methodname{} configuration uses 4.53 scheduled steps on average (5.09 NFE) and requires \textbf{32.7 ms}, an \textbf{8.5$\times$ reduction} relative to $N=50$, with an 88.7\% success rate. Relative to Fixed-Euler $N=4$, \methodname{} adds 6.6 ms of action-head latency (32.7 versus 26.1 ms) while improving success by 5.5 percentage points (88.7\% versus 83.2\%). For reference, fixed Euler with $N=10$ reports 89.0\% success and 54.5 ms, while $N=20$ reports 92.3\% and 109.3 ms. As detailed in Sec.~\ref{sec:sensitivity}, reducing the directional-deviation tolerance to $\epsilon=0.002$ increases the average schedule to 14.1 steps and yields 92.0\% success at 104.4 ms.

\begin{table}[t]
\caption{Action-head latency and success rate on LIBERO. Fixed-threshold and baseline rows aggregate 5 seeds; the OSC row aggregates 3 seeds. For \methodname{} rows, the parenthesized value is the measured average NFE; all probe evaluations are included in action-head latency. Visual encoding is excluded because \methodname{} does not modify the visual encoder.}
\label{tab:libero_results}
\centering
\resizebox{\columnwidth}{!}{%
\begin{tabular}{lccc}
\toprule
\textbf{Method} & \textbf{Reported Count} & \textbf{Flow Solver (ms)} $\downarrow$ & \textbf{Success Rate (\%)} $\uparrow$ \\
\midrule
Fixed-Euler ($N=50$) & 50.0 & 278.7 & \textbf{92.5} \\
Fixed-Euler ($N=20$) & 20.0 & 109.3 & 92.3 \\
Fixed-Euler ($N=10$) & 10.0 & 54.5  & 89.0 \\
Fixed-Euler ($N=5$)  & 5.0  & 34.0  & 86.9 \\
Fixed-Euler ($N=4$)  & 4.0  & 26.1  & 83.2 \\
RK45                 & 101.2 & 3769.3  & 91.2 \\
AB2 & 10.0 & 66.9 & 88.9 \\
Heun ($N=10$) & 20.0 & 116.2 & 83.3 \\
Heun ($N=25$) & 50.0 & 273.22 & 88.2\\
AdaFlow & 28.9 & 184.18 & 90.2\\
\midrule
\methodname{} (fixed $\epsilon$) & \textbf{4.53 (5.09 NFE)} & \textbf{32.7} & 88.7 \\
\textbf{\methodname{} (OSC)} & 5.55 (6.14 NFE) & 36.8 & 89.25 \\
\bottomrule
\end{tabular}
}%
\vspace{-0.3cm}
\end{table}

\subsection{Analysis of Adaptive Scheduling}

\begin{figure}[t]
    \centering
    \includegraphics[width=\columnwidth]{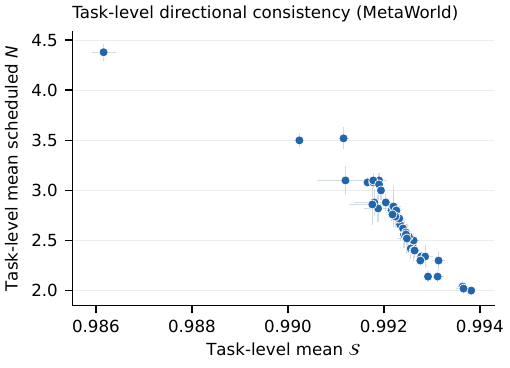}
    \caption{Task-level directional consistency and scheduled integration counts on MetaWorld MT50 at $\Delta t_{\mathrm{probe}}=0.5$, $\epsilon=0.008$, with OSC disabled. Each point represents one task, averaged equally over five seeds with ten episodes per task and seed. Error bars denote standard deviations across seed-level means.}
    \label{fig:directional_consistency}
\end{figure}

Figure~\ref{fig:directional_consistency} shows the distribution of task-level mean $\mathcal{S}$ and the corresponding mean scheduled $N$ under a shared probe configuration. Tasks with higher mean similarity tend to receive fewer integration steps, consistent with the scheduling rule. The spread across tasks illustrates the different average budgets selected under the same threshold. These points summarize task-level averages rather than individual action chunks.

\begin{figure}[t]
    \centering
    \begin{tabular}{p{0.45\columnwidth} p{0.45\columnwidth}}
        \centering \scriptsize \textbf{Baseline: Fixed-Step ($N=50$)} & 
        \centering \scriptsize \textbf{Ours: \methodname{}} \tabularnewline
    \end{tabular}
    
    \vspace{-1mm}
    
    \includegraphics[width=0.95\columnwidth]{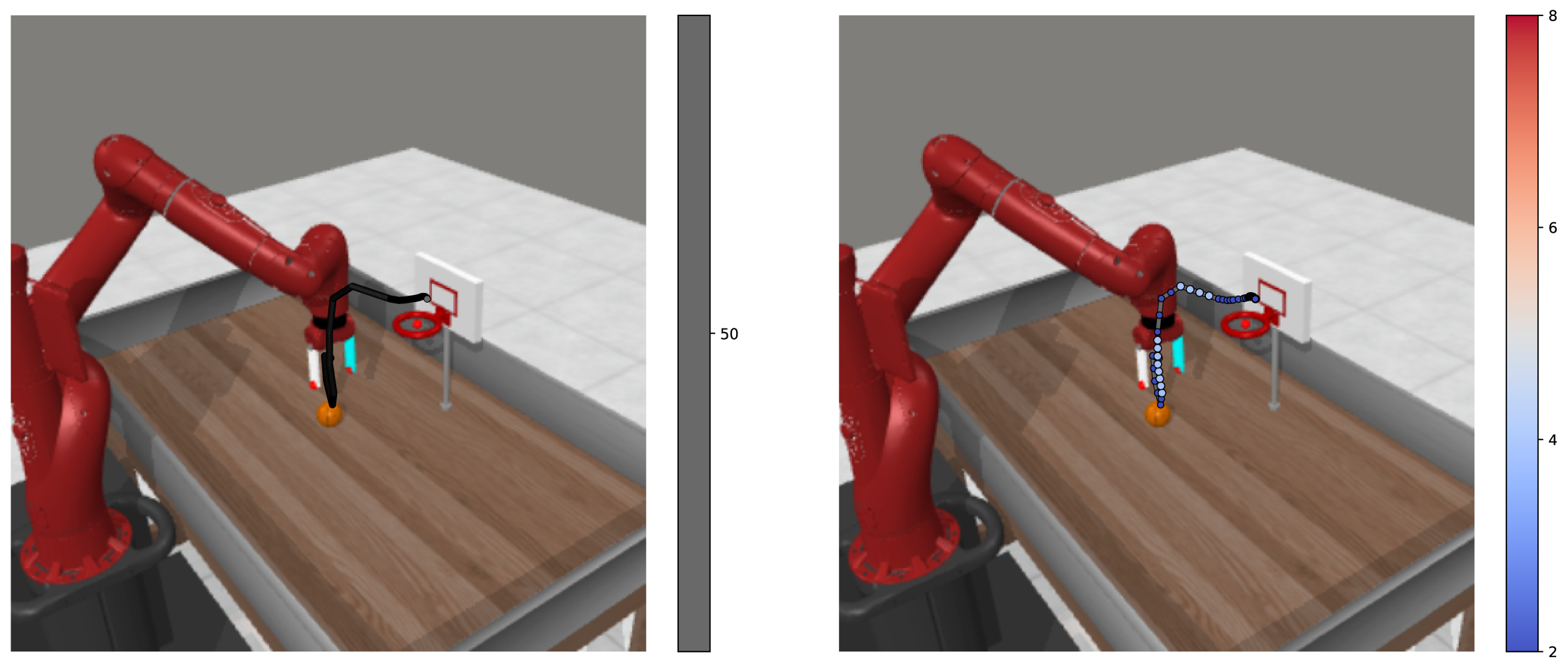}
    
    \vspace{-1mm}
    
    \includegraphics[width=0.95\columnwidth]{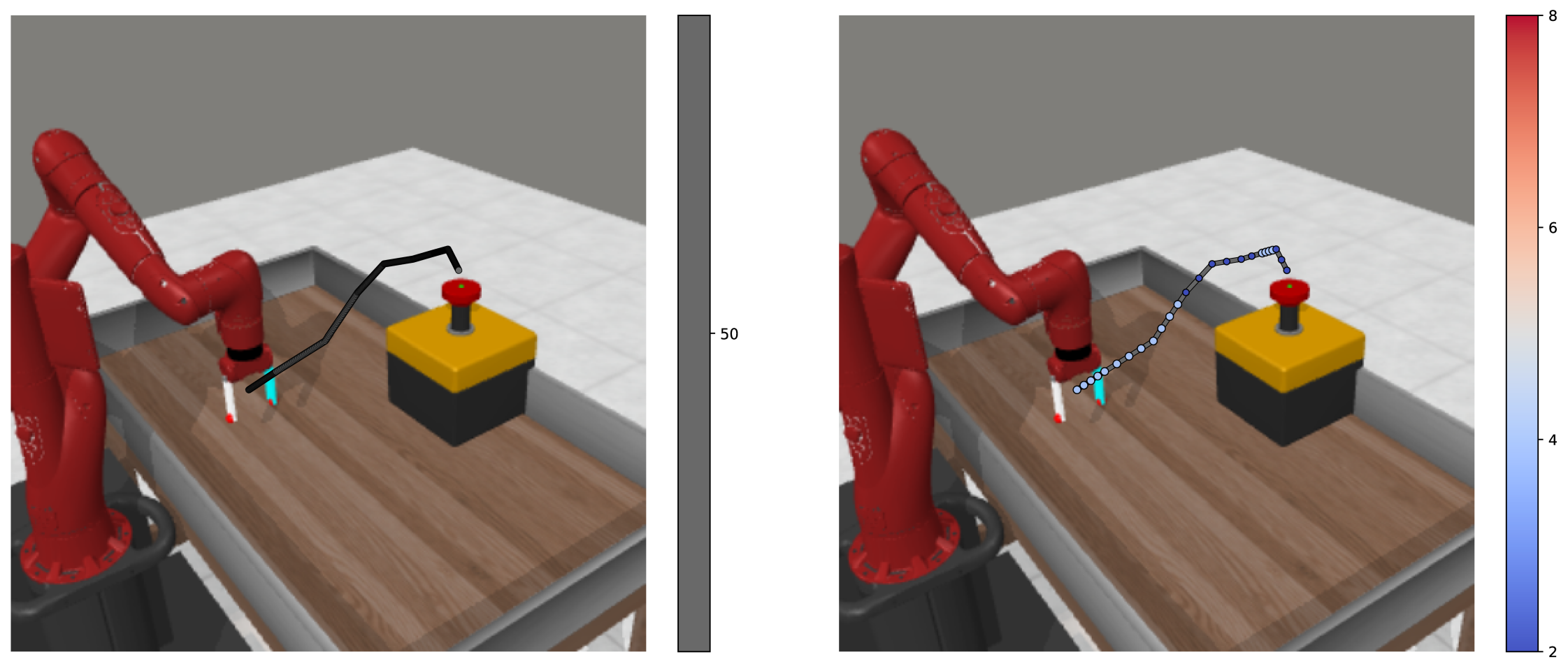}
    
    \vspace{-2mm}
    \caption{Qualitative examples of adaptive step scheduling for two MetaWorld rollouts. \textbf{Top:} In the shown Basketball rollout, the selected schedules contain more steps (red) around object interaction and insertion. \textbf{Bottom:} In the shown Button Press rollout, the selected schedules often use the minimum step count (blue) during transit. Colors visualize the selected schedules.}
    \label{fig:trajectory}
\end{figure}

Figure~\ref{fig:trajectory} visualizes the schedules selected by the probe during two representative rollouts. In the shown Button Press rollout, high cosine similarity ($\mathcal{S} \approx 1.0$) coincides with repeated selection of the minimum schedule ($N=N_{\min}$), which reuses the two probe evaluations in the final update. In the shown Basketball rollout, the scheduler selects more steps around portions involving object interaction and insertion. These examples illustrate the routing behavior of the directional-consistency signal.

\subsection{Oscillation-Conditioned Threshold Adaptation}
\label{sec:osc_results}

We evaluate OSC using the configuration in Sec.~\ref{sec:oscillation_conditioned}, with the probe horizon and policy checkpoints held fixed.

\begin{table*}[!t]
\caption{Fixed-step reference versus OSC with $r_{\mathrm{ref}}=0.02$. OSC results aggregate 3 seeds; the Evo-1 fixed-step references are the 5-seed results from Tables~\ref{tab:metaworld_results} and~\ref{tab:libero_results}. Differences are descriptive comparisons, not paired estimates of the effect of enabling OSC. `--' denotes an unavailable measurement. `OSC Avg. Integration $N$' is reported only when the implementation exposes an adaptive integration schedule; fixed denoising-loop lengths are not relabeled as adaptive steps. For Evo-1, $N$ is a scheduled integration count and the corresponding action-head NFE follows the convention stated in Sec.~\ref{sec:main_results}. The OpenPI/PI0.5 LIBERO-90 result uses 900 episodes per seed; suite-level results use the corresponding evaluation episodes.}
\label{tab:osc_results}
\centering
\setlength{\tabcolsep}{4pt}
\scriptsize
\resizebox{\textwidth}{!}{%
\begin{tabular}{l l c c c c c}
\toprule
\textbf{Model} & \textbf{Benchmark} & \textbf{Fixed Steps} & \textbf{Fixed Success} & \textbf{OSC Success} & \textbf{$\Delta$ Success} & \textbf{OSC Avg. $N$} \\
\midrule
Evo-1 & MetaWorld & 10 & 81.6\% & 84.0\% & +2.4 pp & 3.84 \\
Evo-1 & LIBERO & 10 & 89.0\% & 89.25\% & +0.25 pp & 5.55 \\
OpenPI / PI0.5 & LIBERO-90 & 10 & 27.97\% & 28.26\% & +0.29 pp & 2.32 \\
OpenPI / PI0.5 & LIBERO-object & 10 & 100.0\% & 99.0\% & -1.0 pp & 2.02 \\
OpenPI / PI0.5 & LIBERO-goal & 10 & 97.0\% & 96.33\% & -0.67 pp & 2.02 \\
OpenPI / PI0.5 & LIBERO-10 & 10 & 91.0\% & 94.0\% & +3.0 pp & 2.04 \\
OpenPI / PI0 & ALOHA sim & 10 & 66.7\% & 66.7\% & 0.0 pp & 2.03 \\
SmolVLA & LIBERO-Spatial & 10 & 90.7\% & 93.0\% & +2.3 pp & 5.68 \\
GR00T N1.5 & LIBERO-Spatial & 10 & 96.0\% & 94.0\% & -2.0 pp & 3.04 \\
\bottomrule
\end{tabular}
}%
\end{table*}

Table~\ref{tab:osc_results} summarizes the additional three-seed results. On the Evo-1 action head, OSC achieves 84.0\% success on MetaWorld with an average scheduled integration count of 3.84 (4.21 NFE). On LIBERO, the OSC run achieves 89.25\% over 400 episodes with an average scheduled $N$ of 5.55 (6.14 NFE). The additional NFE arises when OSC routes more chunks to non-minimum schedules; the OSC statistics themselves reuse the probe velocities.

We also test the same sign-change-conditioned threshold mechanism on other VLA policies. For OpenPI/PI0.5, the suite-level results in Table~\ref{tab:osc_results} show the trade-off between step count and success rate: the LIBERO-90 result changes from $27.97\%$ to $28.26\%$, while the suite-level rows include both positive and negative changes. PI0 on ALOHA sim, SmolVLA, and GR00T N1.5 provide additional evaluation settings.

For Evo-1, the fixed 10-step reference and OSC report 81.6\% and 84.0\% success on MetaWorld, and 89.0\% and 89.25\% on LIBERO, respectively. These are separate five-seed reference and three-seed OSC results. The remaining rows show that the observed change depends on the policy and task suite: OpenPI/PI0.5 increases on LIBERO-90 and LIBERO-10, while its LIBERO-object and LIBERO-goal rows decrease. SmolVLA increases from 90.7\% to 93.0\% on LIBERO-Spatial, whereas GR00T N1.5 changes from 96.0\% to 94.0\% on the same suite. These row-wise results characterize the operating points produced by the shared OSC computation across the listed implementations.

\subsection{Sensitivity Analysis}
\label{sec:sensitivity}

A practical requirement for deployment is to avoid additional training and extensive task-by-task tuning. \methodname{} does not introduce learned scheduler parameters. The bounds $N_{\min}$ and $N_{\max}$ specify the permitted integration budget. In our implementation, $N_{\min}=2$ corresponds to the two action-head evaluations used by the probe-based minimum schedule.

\begin{table}[t]
\caption{Sensitivity analysis of the directional-deviation tolerance $\epsilon$ across MetaWorld and LIBERO, aggregated over 5 seeds with 10 episodes per task and seed. `Avg. Scheduled $N$' is the selected integration schedule; measured action-head latency includes the probe evaluation.}
\label{tab:ablation_epsilon}
\centering
\resizebox{\columnwidth}{!}{%
\begin{tabular}{llccc}
\toprule
\textbf{Domain} & \textbf{Threshold} $\epsilon$ & \textbf{Avg. Scheduled $N$} & \textbf{Action Head (ms)} $\downarrow$ & \textbf{Success Rate (\%)} \\
\midrule
MetaWorld & 0.001 & 15.7 & 90.6 & 82.8 \\
MetaWorld & 0.002 & 8.6  & 47.3  & 81.1 \\
MetaWorld & 0.004 & 4.6  & 28.4  & 82.7 \\
MetaWorld & 0.008 & \textbf{2.6}  & \textbf{15.9}  & \textbf{83.2} \\
LIBERO & 0.001 & 19.7 & 135.1 & 91.6 \\
LIBERO & 0.002 & 14.1 & 104.4 & \textbf{92.0} \\
LIBERO & 0.004 & 7.9  & 61.0  & 90.3 \\
LIBERO & 0.008 & \textbf{4.5}  & \textbf{32.7}  & 88.7 \\
\bottomrule
\end{tabular}
}%
\end{table}

Table~\ref{tab:ablation_epsilon} reports the measured compute--performance trade-off for $\epsilon \in \{0.001, 0.002, 0.004, 0.008\}$. On MetaWorld, the largest tested tolerance, $\epsilon=0.008$, yields the lowest measured action-head latency (15.9 ms) and the highest measured success rate in this sweep (83.2\%).

On LIBERO, $\epsilon=0.008$ yields 4.5 average steps and 88.7\% success. Reducing the tolerance to $\epsilon=0.002$ increases the average schedule to 14.1 steps and yields 92.0\% success, compared with 92.5\% for fixed Euler $N=50$, while retaining a lower measured action-head latency (104.4 ms versus 278.7 ms).

The selected tolerance changes the compute--performance operating point. \methodname{} provides a controllable, training-free acceleration mechanism, and the tolerance can be selected according to a deployment latency target and the required empirical performance level.

\begin{table*}[!t]
\caption{Multi-task real-world evaluation with 50 trials per method and task. Success rates and 95\% Wilson intervals are computed from the number of successful trials out of 50. The reported steps are averaged across episodes.}
\label{tab:real_world_results}
\centering
\setlength{\tabcolsep}{3pt}
\scriptsize
\resizebox{\textwidth}{!}{%
\begin{tabular}{llcccccc}
\toprule
\textbf{Task} & \textbf{Method} & \textbf{Success} & \textbf{95\% Wilson CI} & \textbf{Solver (ms)} & \textbf{Steps} & \textbf{Obs. $\Delta$ (normalized)} & \textbf{Obs. jerk (normalized)} \\
\midrule
pick-place fruit & $\pi_{0.5}$ & 82\% & 69.2--90.2\% & 183.9 & 10.0 & 0.06 & 0.03 \\
pick-place fruit & $\pi_{0.5}$  & 84\% & 71.5--91.7\% & 354.7 & 20.0 & 0.055 & 0.025 \\
pick-place fruit & $\pi_{0.5}$  & 84\% & 71.5--91.7\% & 885.0 & 50.0 & 0.050 & 0.020 \\
pick-place fruit & \methodname{} OSC & \textbf{90\%} & 78.6--95.7\% & \textbf{59.6} & $\mathbf{2.6}$ & 0.07 & 0.04 \\
flip mug & $\pi_{0.5}$ & 52\% & 38.5--65.2\% & 185.3 & 10.0 & 0.04 & 0.03 \\
flip mug & $\pi_{0.5}$ & 54\% & 40.4--67.0\% & 357.0 & 20.0 & 0.035 & 0.025 \\
flip mug & $\pi_{0.5}$  & 54\% & 40.4--67.0\% & 880.0 & 50.0 & 0.030 & 0.020 \\
flip mug & \methodname{} OSC & \textbf{54\%} & 40.4--67.0\% & \textbf{68.4} & $\mathbf{3.3}$ & 0.05 & 0.02 \\
fold/stack cups & $\pi_{0.5}$ & 38\% & 25.9--51.8\% & 184.7 & 10.0 & 0.05 & 0.06 \\
fold/stack cups & $\pi_{0.5}$  & 40\% & 27.6--53.8\% & 356.0 & 20.0 & 0.045 & 0.050 \\
fold/stack cups & $\pi_{0.5}$ & 42\% & 29.4--55.8\% & 878.0 & 50.0 & 0.040 & 0.040 \\
fold/stack cups & \methodname{} OSC & \textbf{46\%} & 33.0--59.6\% & \textbf{125.8} & $\mathbf{7.4}$ & 0.06 & 0.04 \\
\bottomrule
\end{tabular}
}%
\end{table*}

\begin{figure}[t]
    \centering
    \begin{tabular}{c}
        \includegraphics[width=0.95\columnwidth]{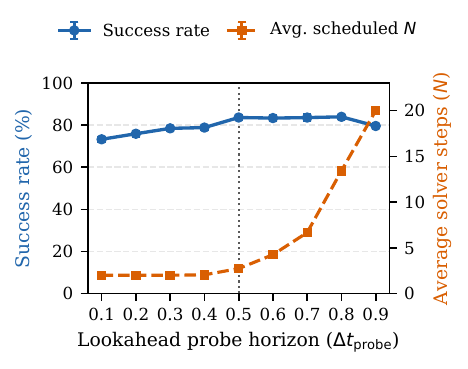}
    \end{tabular}
    \caption{Probe-horizon ablation on MetaWorld (MT50). Points show means over five seeds; error bars denote standard deviations. The average scheduled integration count (dashed orange) generally increases with the probe horizon, while success-rate means (solid blue) range from 83.2\% to 84.0\% for horizons from 0.5 to 0.8.}
    \label{fig:probe_ablation}
\end{figure}

\noindent \textbf{Ablation on Lookahead Probe Horizon.}
We sweep $\Delta t_{\mathrm{probe}}\in\{0.1,\ldots,0.9\}$ on MetaWorld MT50 while fixing the checkpoint, routing threshold, and all other solver settings. Figure~\ref{fig:probe_ablation} aggregates five seeds, each with ten episodes per task. Short horizons ($0.1$--$0.4$) select the minimum two-step schedule almost exclusively ($N=2.00$--$2.05$) and report 73.3\%--78.9\% mean success. At $\Delta t_{\mathrm{probe}}=0.5$, this ablation reports $83.2\%$ success and $2.60$ scheduled steps.

For horizons from $0.5$ to $0.8$, mean success rates range from $83.2\%$ to $84.0\%$, while the average scheduled count rises from $2.60$ to $13.42$. At $0.9$, the scheduler reaches its maximum budget ($N=20$) and reports $79.6\%$ mean success. We use $\Delta t_{\mathrm{probe}}=0.5$ as the default because it has the lowest average scheduled count among the settings from $0.5$ to $0.8$. 

\subsection{Real-World Deployment}
\label{sec:real_world}

We evaluate \methodname{} in a real-world setting to measure action-head latency, selected schedules, and task outcomes under a fixed hardware configuration. The setup consists of a 7-DoF UFACTORY xArm7 equipped with a 6-DoF Inspire multi-finger dexterous hand. Visual observations are captured via a fixed wrist-mounted Orbbec Gemini 360 camera, providing a consistent egocentric perspective for the VLA policy. We evaluate three real-world manipulation tasks: ``pick-place fruit'', ``flip mug'', and ``fold/stack cups''. These tasks involve grasp alignment, object reorientation, and sequential placement, respectively.

\begin{figure}[t]
    \centering
    \setlength{\tabcolsep}{2pt}
    \renewcommand{\arraystretch}{0.8}
    \resizebox{\columnwidth}{!}{%
    \begin{tabular}{ccccc}
        \multicolumn{5}{c}{\small\textbf{Pick-place fruit}} \\[2mm]
        \includegraphics[width=0.18\columnwidth]{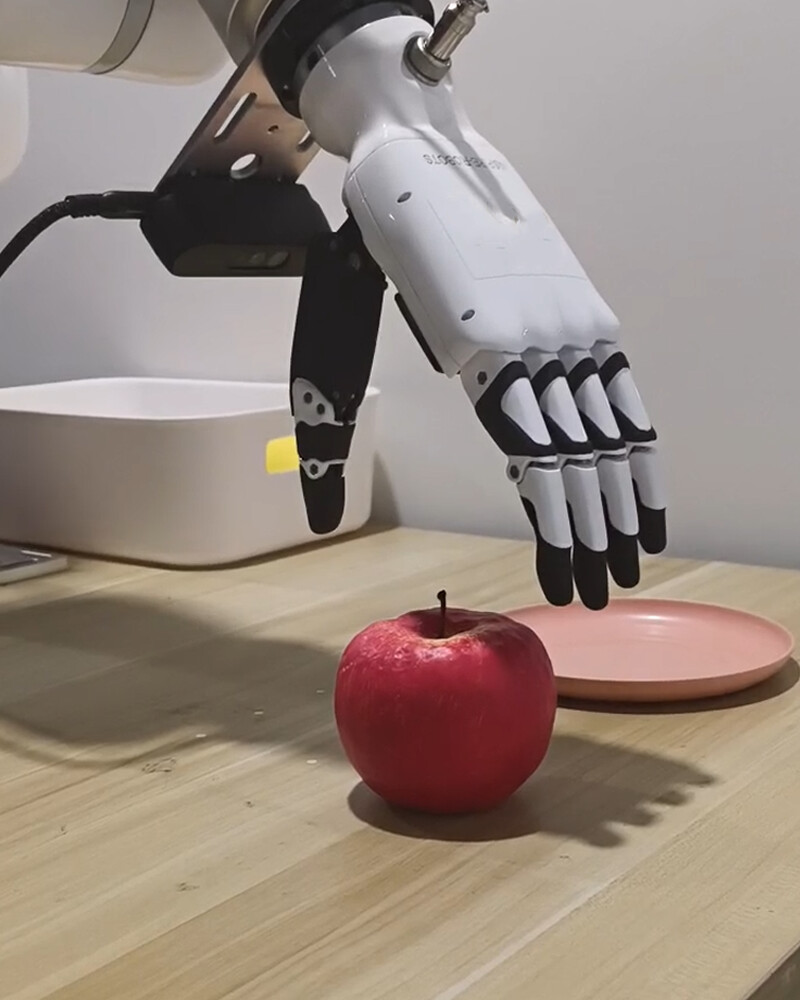} &
        \includegraphics[width=0.18\columnwidth]{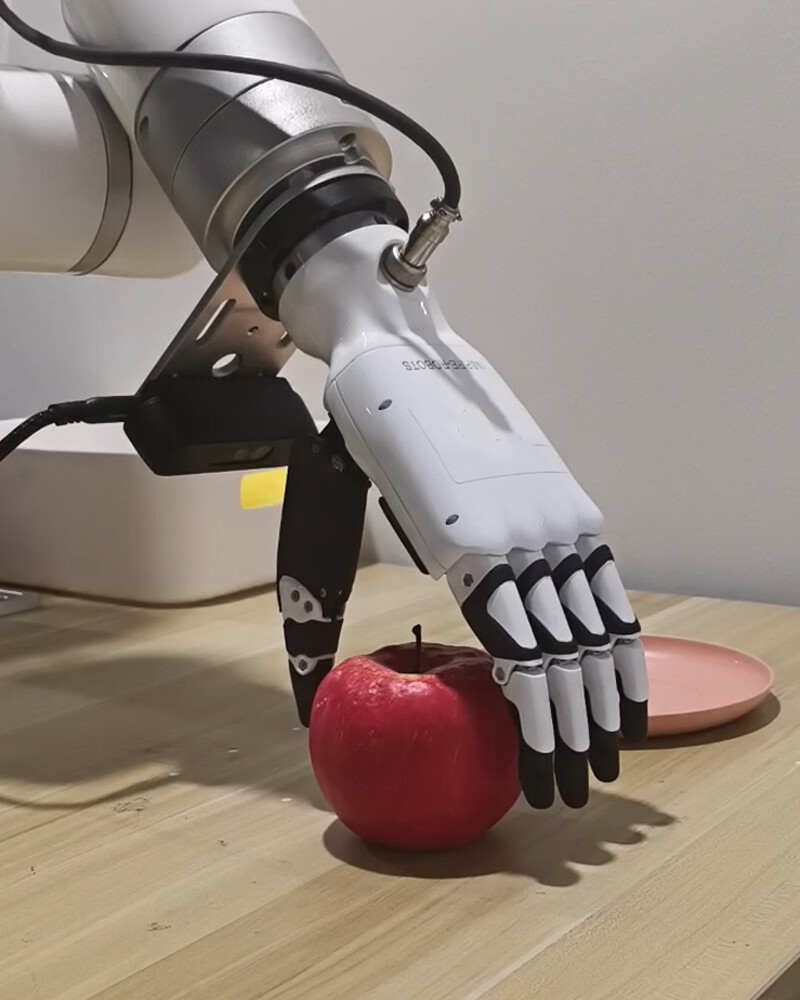} &
        \includegraphics[width=0.18\columnwidth]{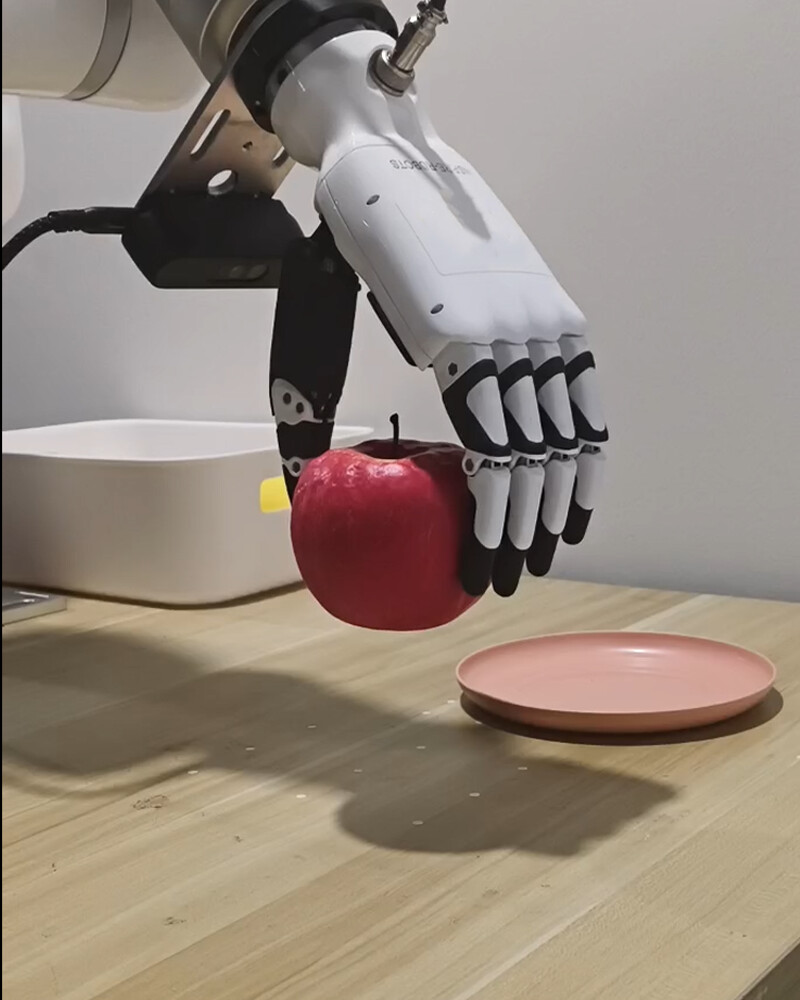} &
        \includegraphics[width=0.18\columnwidth]{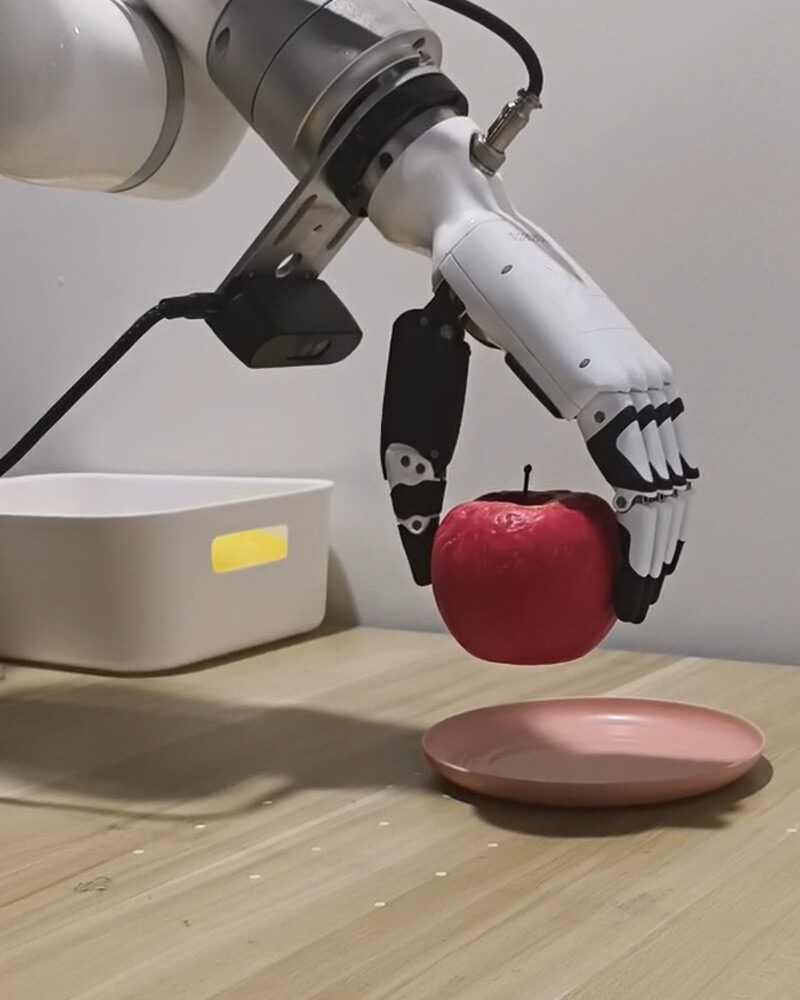} &
        \includegraphics[width=0.18\columnwidth]{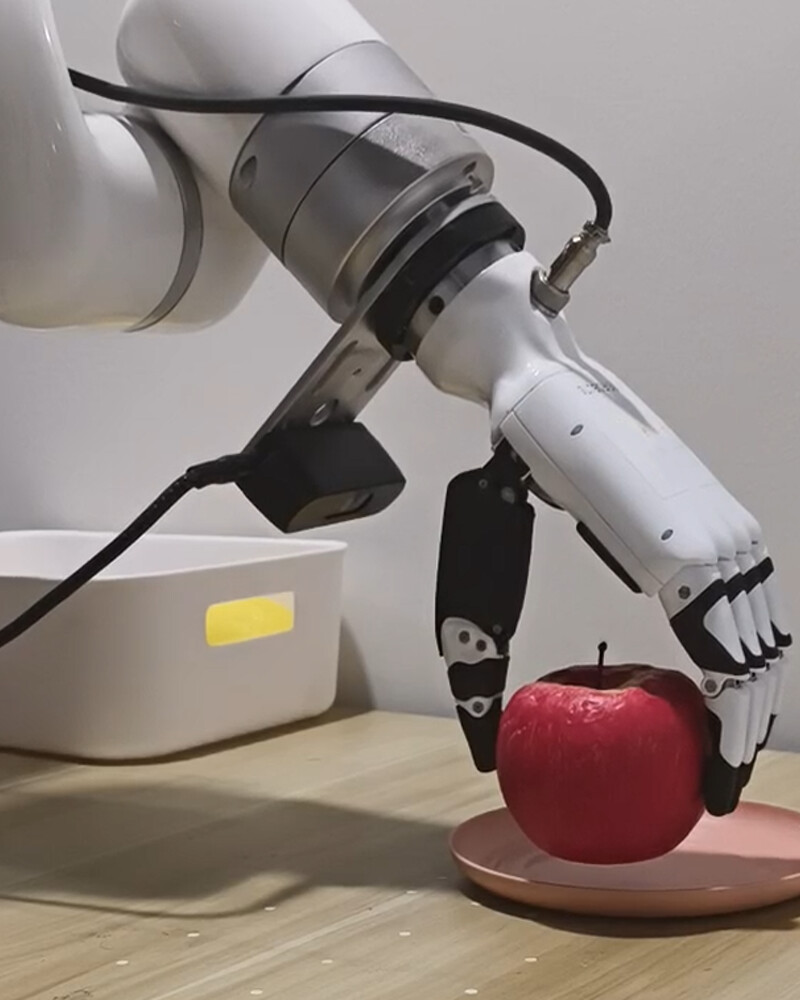} \\[2mm]

        \multicolumn{5}{c}{\small\textbf{Flip mug}} \\[2mm]
        \includegraphics[width=0.18\columnwidth]{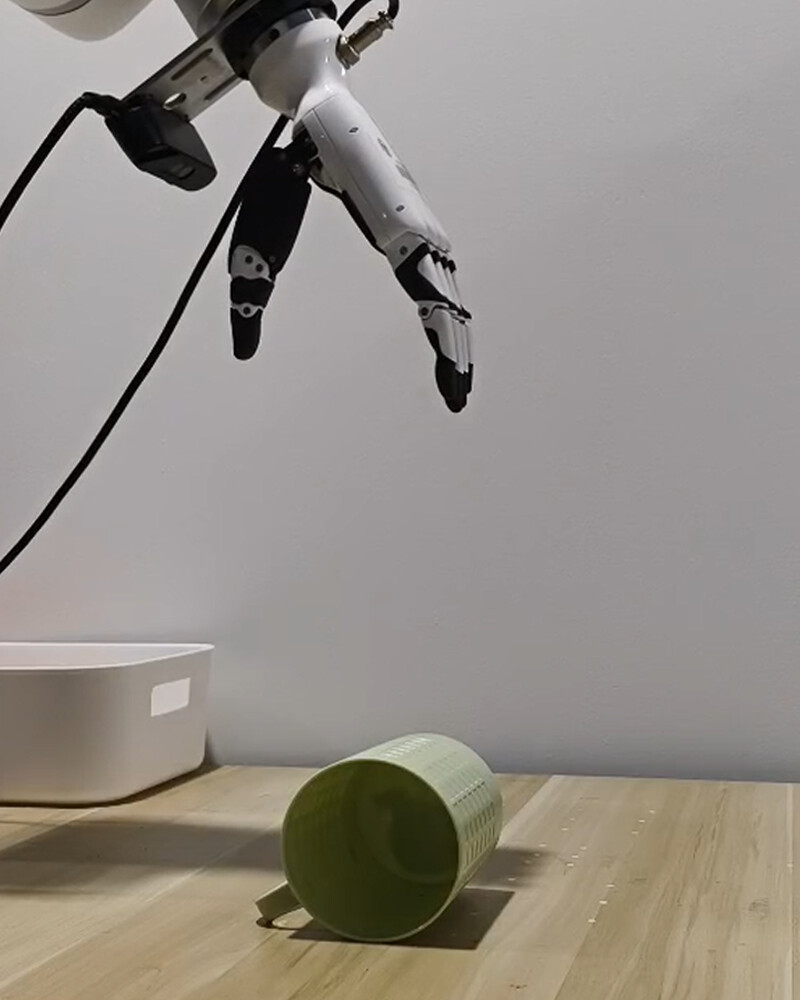} &
        \includegraphics[width=0.18\columnwidth]{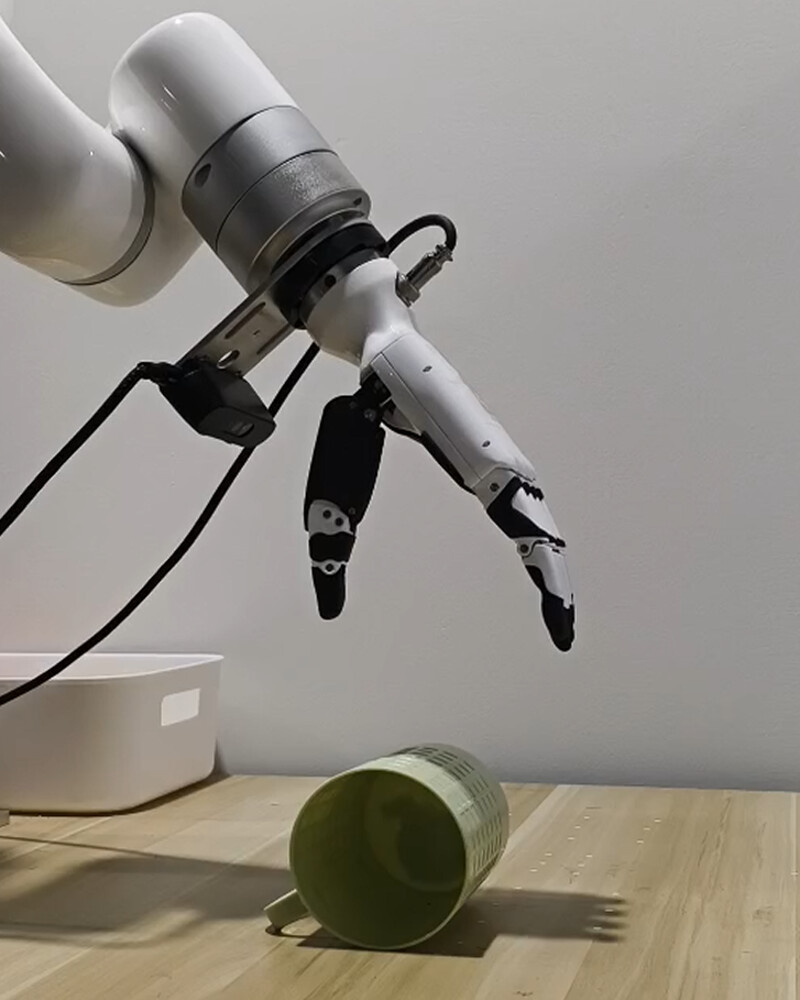} &
        \includegraphics[width=0.18\columnwidth]{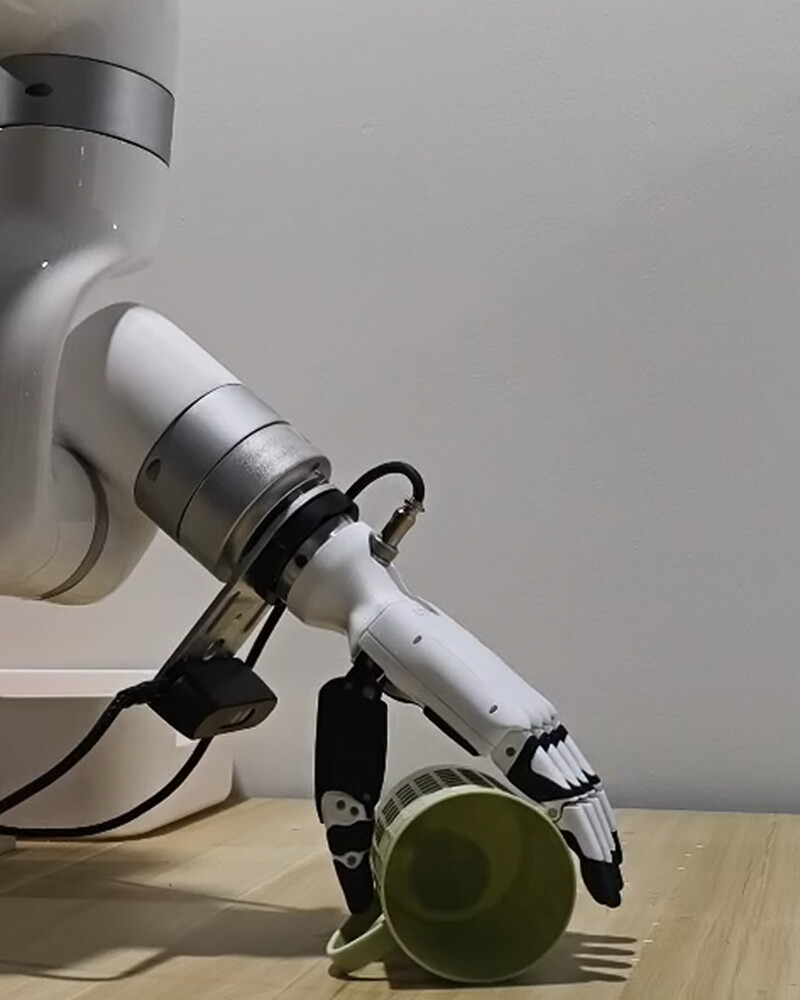} &
        \includegraphics[width=0.18\columnwidth]{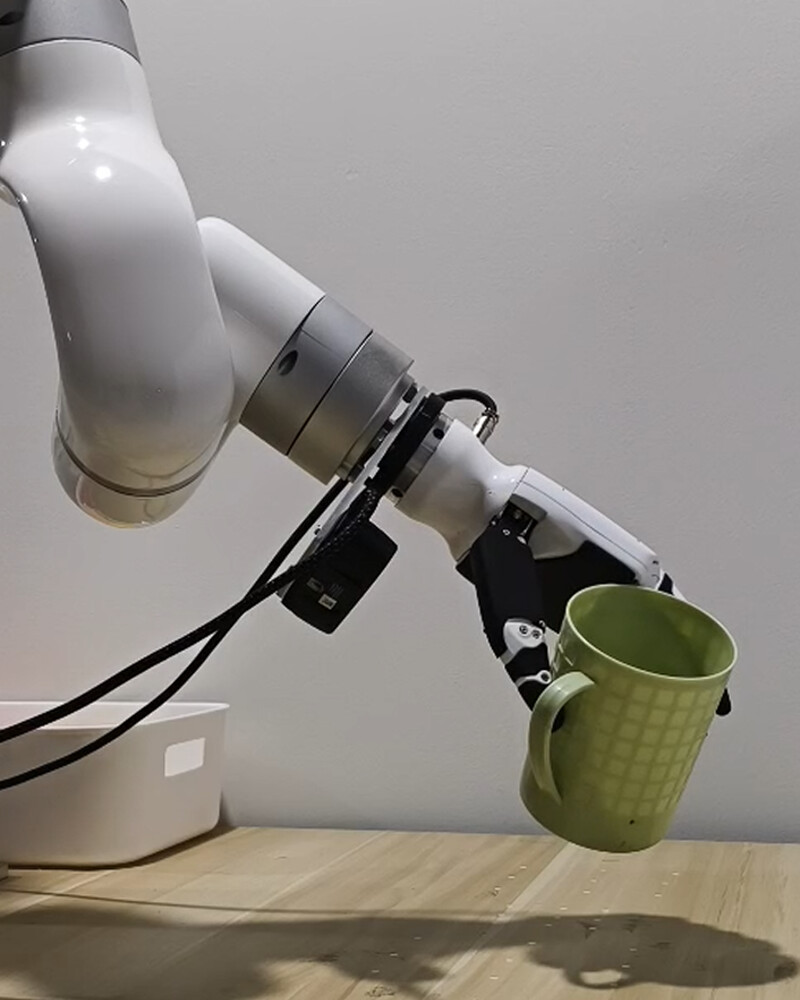} &
        \includegraphics[width=0.18\columnwidth]{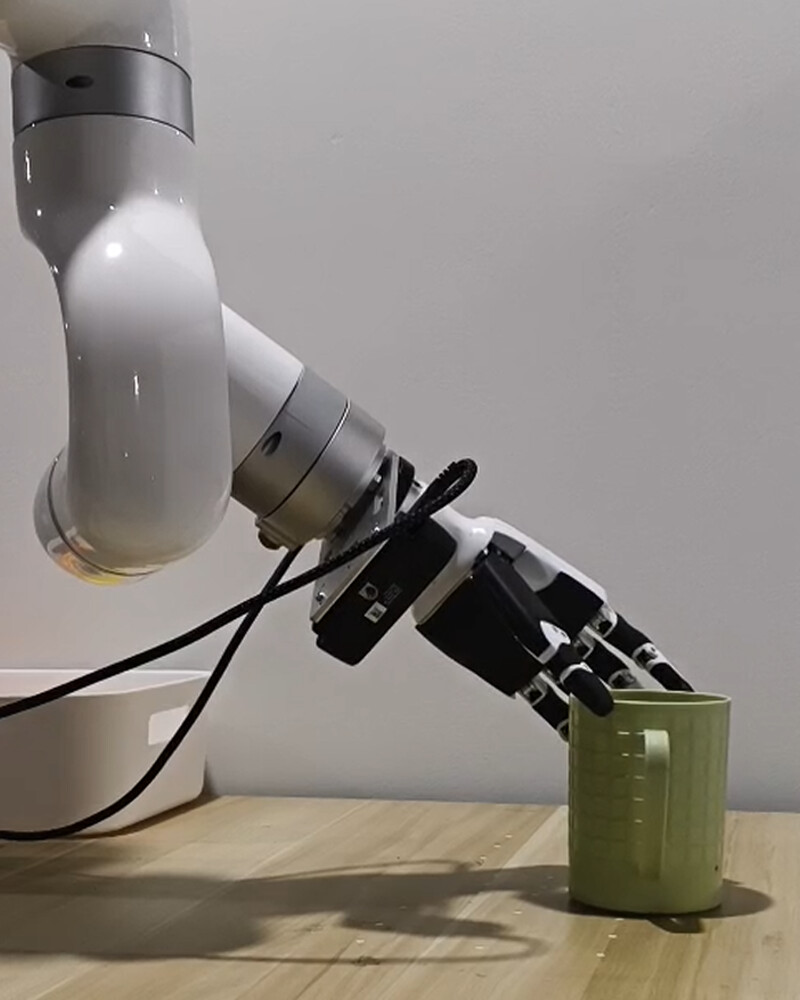} \\[2mm]

        \multicolumn{5}{c}{\small\textbf{Fold/stack cups}} \\[2mm]
        \includegraphics[width=0.18\columnwidth]{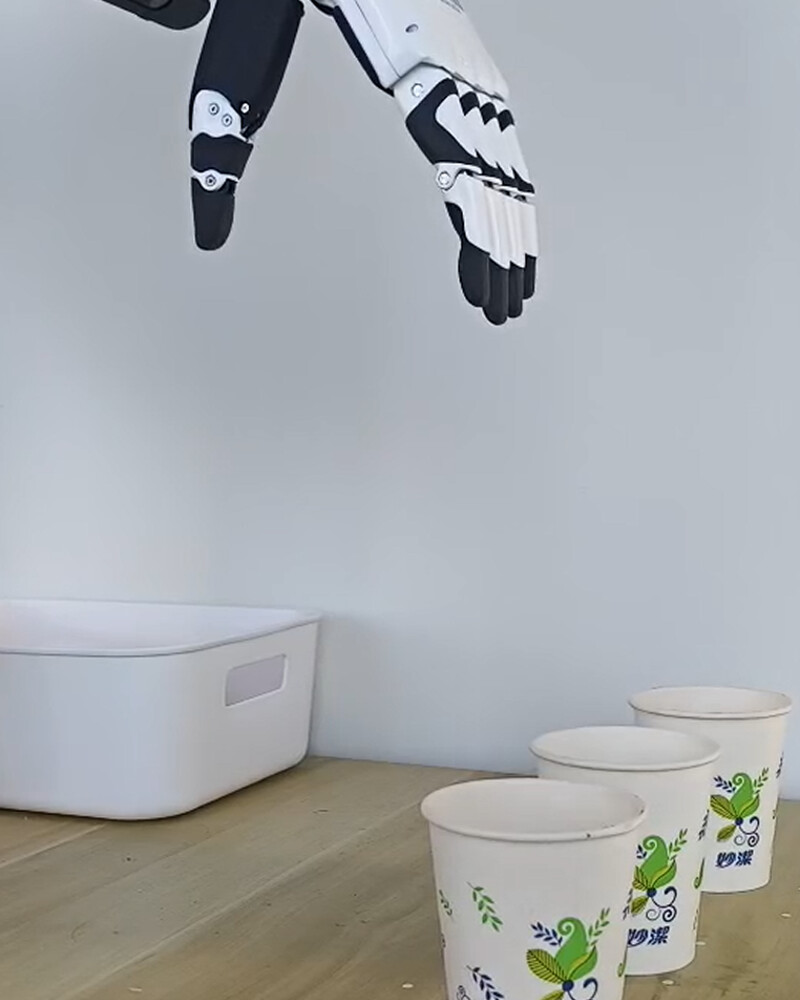} &
        \includegraphics[width=0.18\columnwidth]{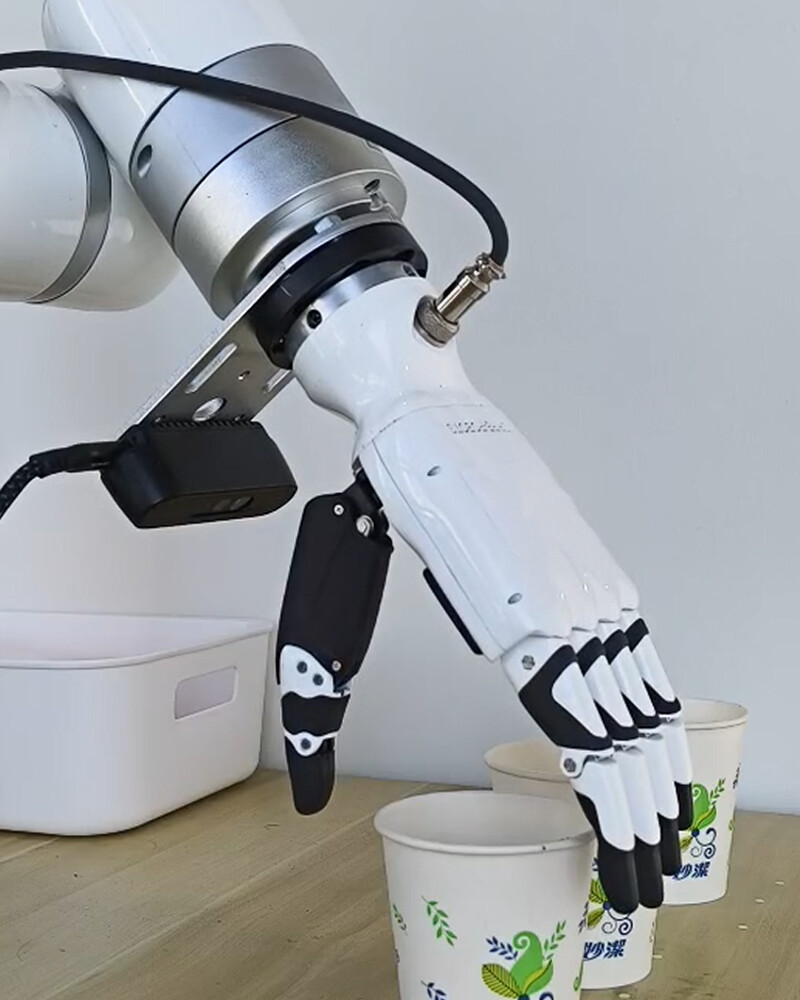} &
        \includegraphics[width=0.18\columnwidth]{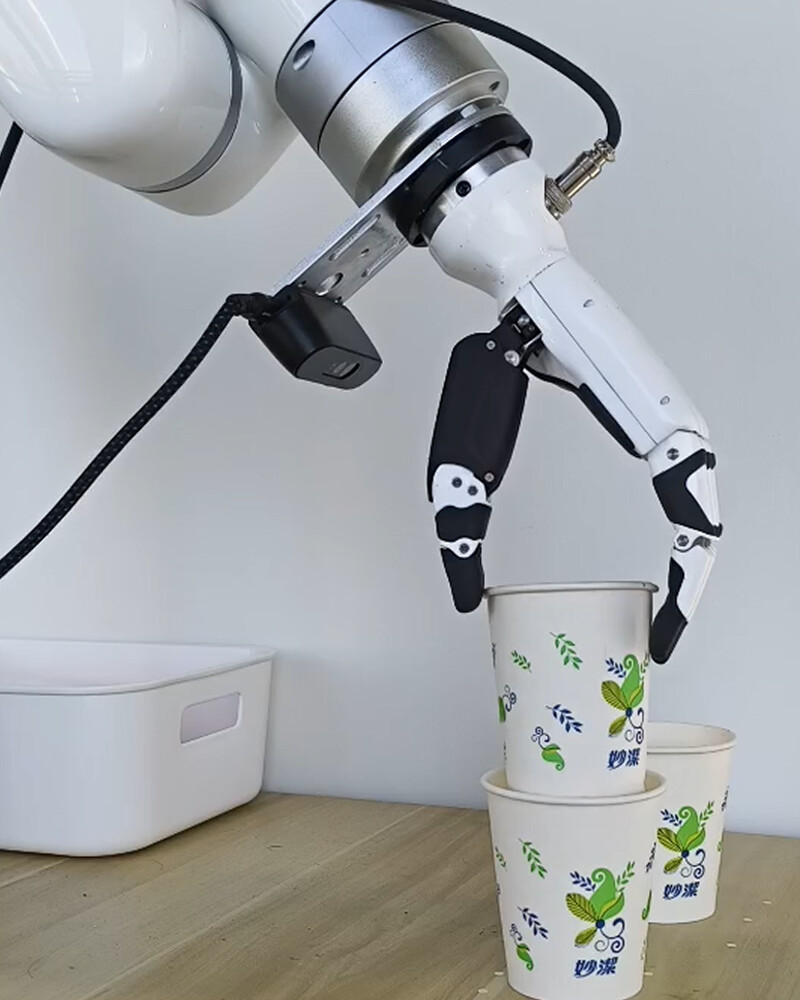} &
        \includegraphics[width=0.18\columnwidth]{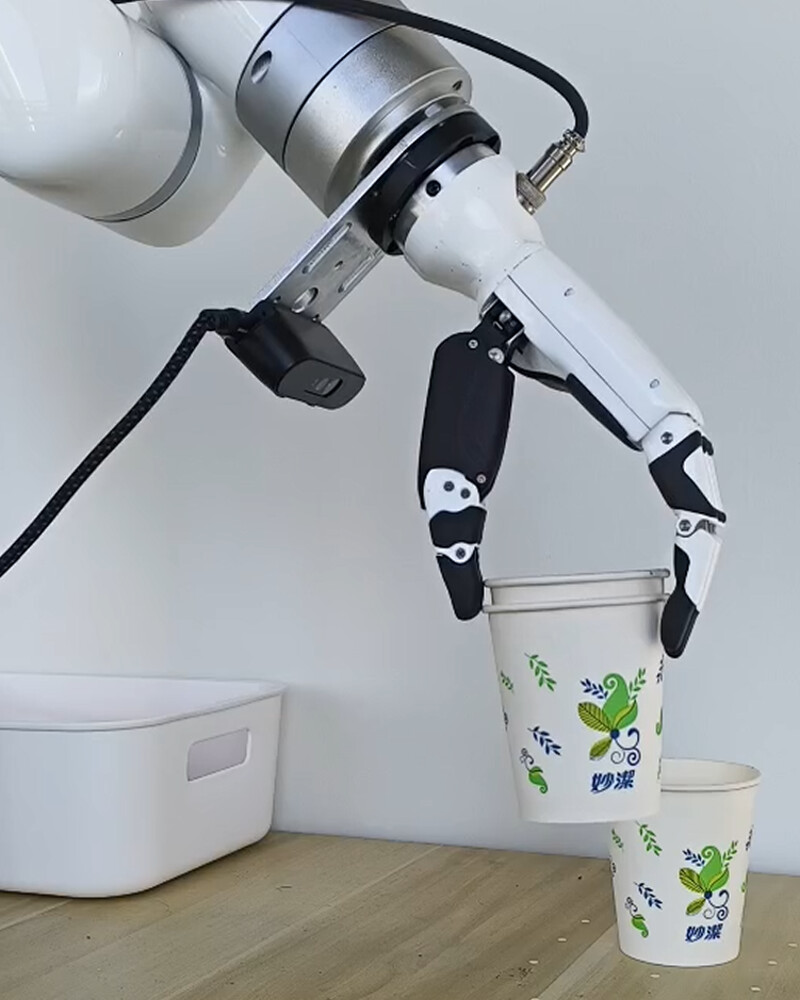} &
        \includegraphics[width=0.18\columnwidth]{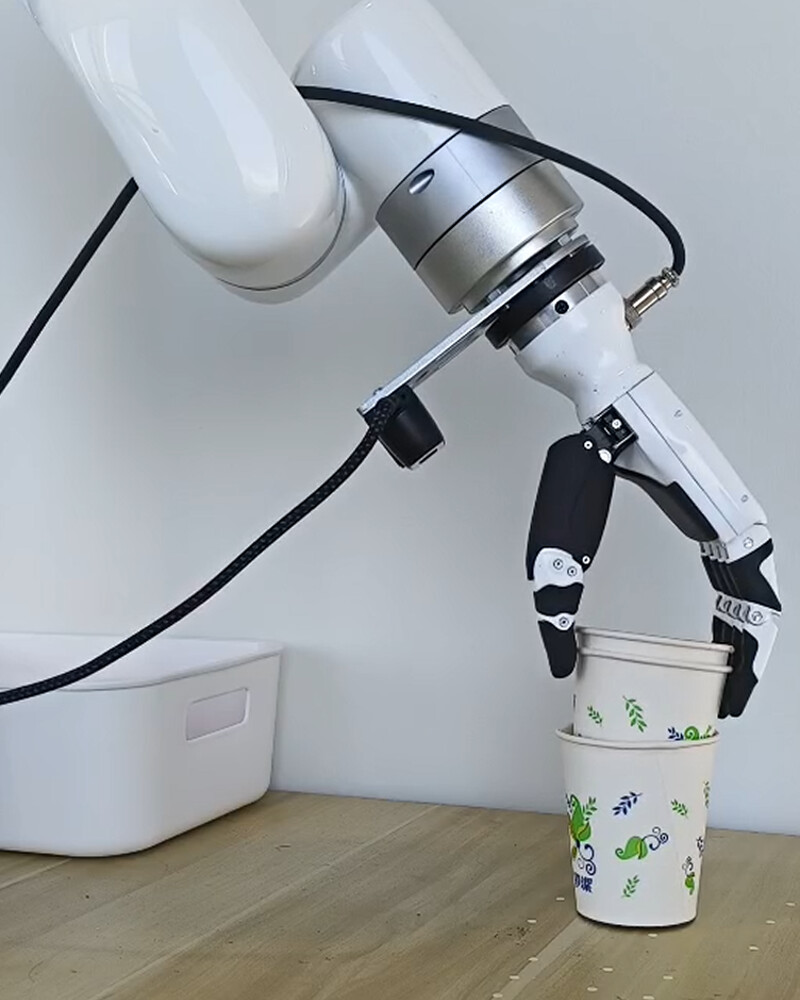}
    \end{tabular}%
    }
    \caption{Representative real-world manipulation sequences. Each row shows the temporal progression of one task under \methodname{}.}
    \label{fig:real_world}
\end{figure}

Figure~\ref{fig:real_world} shows representative executions with the \methodname{} OSC action head. In Table~\ref{tab:real_world_results}, \methodname{} has lower reported solver latency and fewer average steps than the listed $\pi_{0.5}$ settings for all three tasks. Its observed success rate is 90\%, 54\%, and 46\% for pick-place fruit, flip mug, and fold/stack cups, respectively; the Wilson intervals show the uncertainty associated with the available trials. The observed joint-displacement and jerk statistics are numerically larger for \methodname{} in these trials, while the direction of the jerk difference varies by task.

Because the visual encoder and communication path are outside the scope of \methodname{}, we report solver latency separately; these measurements characterize the evaluated action head rather than end-to-end latency or robustness. The quasi-static tasks do not establish benefits for dynamic-target tracking or disturbance recovery.

\noindent \textbf{Qualitative Failure Analysis.} We use a lightweight taxonomy based on the first observable failure phase and failure manifestation. For pick-place fruit, observed failures include grasp misalignment and object slippage after lifting. For mug flipping, failures include contact between the fingers and the mug that triggers a mechanical stop, as well as grasped mugs that are not successfully reoriented. For fold/stack cups, failures include failure to grasp a cup, knocking over cups during stacking, and failure in a later cup after an earlier stacking stage succeeds. These labels describe observable outcomes rather than definitive causal explanations; perception, data quality, grasping, object state, and trajectory execution may contribute jointly. We did not observe a clearly \methodname{}-specific failure category in the evaluated trials.

\section{CONCLUSION}

We propose \methodname{}, a training-free adaptive solver for Flow-Matching VLA action heads. Its Lookahead Consistency Probe compares initial and lookahead velocity directions and uses the resulting local directional-consistency signal to select an integration schedule. The optional OSC mechanism modulates the routing threshold online to expose an additional accuracy--compute trade-off. Simulation and real-world evaluations characterize the action-head latency--success trade-offs of \methodname{} relative to fixed-step baselines. The directional-deviation tolerance can be selected for the target benchmark and latency requirement. Future work should evaluate the routing signal under more dynamic, contact-rich physical tasks and determine whether the fixed probe horizon can be adapted reliably across policies and deployment conditions.
\balance
\bibliographystyle{IEEEtran}
\bibliography{references}

\end{document}